\documentclass[11pt,a4paper]{article}
\usepackage[hyperref]{acl2021}
\usepackage{times}
\usepackage{latexsym}

\usepackage{amsfonts,amsmath}
\usepackage{url}
\usepackage{paralist}
\usepackage{amsmath}
\usepackage{multirow}
\usepackage{array}
\usepackage{multicol}
\usepackage{caption}
\usepackage{enumerate}
\usepackage{graphicx}
\usepackage{enumitem}
\usepackage{bm}
\usepackage{url}
\usepackage[normalem]{ulem}
\usepackage{subfigure}
\usepackage{mathrsfs}
\usepackage{ulem}
\usepackage{booktabs}
\usepackage{metalogo}

\newcommand{\ffn}{\mathop{\mathrm{FFN}}}
\newcommand{\selfatt}{\mathop{\mathrm{Self\_Attention}}}
\newcommand{\embed}{\mathop{\mathrm{Embed}}}
\newcommand{\gru}{\mathop{\mathrm{GRU}}}

\usepackage{microtype}

\aclfinalcopy % Uncomment this line for the final submission
\newcommand\blfootnote[1]{%
  \begingroup
  \renewcommand\thefootnote{}\footnote{#1}%
  \addtocounter{footnote}{-1}%
  \endgroup
}

\title{Quotation Recommendation and Interpretation \\ Based on Transformation from Queries to Quotations}

\author{Lingzhi Wang$^{1,2}$, Xingshan Zeng$^{3}$, Kam-Fai Wong$^{1,2}$\\
  $^1$The Chinese University of Hong Kong, Hong Kong, China\\
  $^2$MoE Key Laboratory of High Confidence Software Technologies, China\\
  $^3$Huawei Noah's Ark Lab, China \\
  \tt $^{1,2}$\{lzwang,kfwong\}@se.cuhk.edu.hk \tt $^3$zeng.xingshan@huawei.com \\
}

\date{}

\begin{document}
\maketitle

\begin{abstract}
%Quotations are essential for successful persuasion and explanation in interpersonal communication. 
To help individuals express themselves better, quotation recommendation is receiving growing attention. 
Nevertheless, most prior efforts focus on modeling quotations and queries separately and ignore the relationship between the quotations and the queries. 
In this work, we introduce a transformation matrix that directly maps the query representations to quotation representations. 
To better learn the mapping relationship, we employ a mapping loss that minimizes the distance of two semantic spaces (one for quotation and another for mapped-query). 
Furthermore, we explore using the words in history queries to interpret the figurative language of quotations, where quotation-aware attention is applied on top of history queries to highlight the indicator words. 
Experiments on two datasets in English and Chinese show that our model outperforms previous state-of-the-art models. 
\end{abstract}
\section{Introduction}
%The prominent use of social media platforms (such as Weibo and Twitter) exposes people with opportunities to share viewpoints and discuss topics they are interested in. 
Quotations are essential for successful persuasion and explanation in interpersonal communication.\blfootnote{The code is available at \url{https://github.com/Lingzhi-WANG/Quotation-Recommendation}}
% Clear, beautiful and persuasive language becomes more and more important in individuals' daily life and online discussions, where quotations can take an important role. 
%Quotations are memorable phrases or sentences widely echoed to spread patterns of wisdom. 
However, it is a daunting task for many individuals to write down a suitable quotation in a short time. This results in a pressing need to develop a quotation recommendation tool to meet such a demand. 

To that end, extensive efforts have been made to \textbf{quotation recommendation}, which aims to recommend an ongoing conversation with a quotation whose sense continues with the existing context~\cite{wang-etal-2020-continuity}. 
%Nevertheless, previous work mostly focused on modeling quotations and contexts separately, largely ignoring exploring the direct relationship between quotations and contexts.  
As quotations are concise phrases or sentences to spread wisdom, which are always in figurative language and difficult to understand, they are assumed written in a different pseudo-language \cite{liu-etal-2019-neural-based}. 
%In this work, we assume the semantic spaces of quotations and queries are
Intuitively, we can infer the meanings of quotations by their neighborhood contexts, especially by the \textit{query} turn (the last turn of conversation that needs recommendation).
%In this work, we assume we can infer the meaning of quotations according to the \textit{query} turn (the last turn of conversation). 
%To capture the relationship between representations of quotations and queries, we introduce a transformation matrix that directly maps the queries to quotations.

\begin{figure}[t]\footnotesize
    \centering
    \begin{tabular}{|p{7.2cm}|}
    \hline
    % \\
    $[t_1]$: Save your money.  Scuf is the biggest ripoff in gaming. \\
    $[t_2]$: What would you suggest instead? \\
    $[t_3]$: Just use a normal controller. \\
    $[t_4]$: Ooooooh, I get it now...you're just dumb. \\
    $[t_5]$: The dumb ones are the people spending over \$100 for a controller. [\textcolor{blue}{\textit{A fool and his money are soon parted.}}] \\
    % ...\\
    \hline
    $[h_1]$: Anyone that \uwave{spends} that \uwave{much} \uwave{money} just to get different writing on a box..... [\textcolor{blue}{\textit{A fool ... parted.}}]\\
    $[h_2]$: And that's probably why you'll \uwave{never} have a \uwave{billion} \uwave{dollars}. [\textcolor{blue}{\textit{A fool ... parted.}}]\\
    $[h_3]$: Seriously. Why do people not do \uwave{market} \uwave{research} \uwave{before} \uwave{buying} something!?! [\textcolor{blue}{\textit{A fool ... parted.}}]\\
    \hline
    \end{tabular}
    % \vskip -0.5em
    \caption{A Reddit conversation snippet (upper part) with three history queries (lower part). Quotations to be recommended are in square brackets. Indicative words are on wavy-underline.
    %[$t_5$] and [$h_1$] to [$h_3$] are query turns. 
    }
    \label{fig:intro-example}
    % \vskip -2.0em
\end{figure}
To illustrate our motivation, Figure \ref{fig:intro-example} shows a Reddit conversation with some history queries associated with quotation $Q$,  ``\textit{A fool and his money are soon parted}". 
% The last turn of the conversation is denoted as \textit{query} turn. 
From the queries ($t_5$ and $h_1$ to $h_3$), we can infer the meaning of quotation $Q$ is ``A foolish person spends money carelessly and won't have a lot of money." based on the contexts.
% $t_5$, $h_1$, and $h_2$.
From $h_3$, we can also know the implication behind the words, which is ``Do a marketing research before buying". 
% Since humans can establish the relationship between quotations and queries, so can machines (neural network). 
Humans can establish such a relationship between quotations and queries and then decide what to quote in their writings, so can machines (neural network).
% Therefore, we introduce a transformation matrix that directly maps the query representations to quotation representations. 
Therefore, we introduce a transformation matrix, in which machines can learn the direct mapping from queries to quotations.
The matrix is worked on the outputs of two encoders, conversation encoder and quotation encoder, encoding conversation context and quotations respectively.

% Furthermore, we explore interpreting the figurative language of quotations based on our main model. 
% %In our writing(either formal writing or online discussions), we generally decide what to quote based on the context that we are currently involved in. 
% Intuitively, some phrases in the context may signal the important information relative to the quotation. To examine whether our model indeed learns useful knowledge and perform recommendation reasonably, we compute quotation-aware attention over the all history queries that are related to the quotation, and then display the most relative queries

Furthermore, we can use the words in the queries to interpret quotations. $h_1$ to $h_3$ in Figure \ref{fig:intro-example} are denoted as \textit{history queries}, and the words on wavy-underline are denoted as indicators to quotations. It can be seen that we can interpret quotations by highlighting the words in the queries. Therefore, we compute quotation-aware attention over all the history queries (after the same transformation as we mentioned before) and then display indicators we learned, which also reflects 
% the transformation working.
the effectiveness of the transformation.

In summary, we introduce a transformation between the query semantic space and quotation semantic space. To minimize the distance of their semantic space after transformation mapping, an auxiliary mapping loss is employed. Besides, we propose a way to interpret quotations with indicative words in the corresponding queries.

%Take quotation $Q$ in Figure \ref{fig:intro-example} as an example. As can be seen, a human can infer the meaning of such quotation from the words with wavy-underline, which are denoted as indicators to quotations.

The remainder of this paper is organized as follows. The related work is surveyed in Section \ref{sec:related_work}. Section \ref{sec:model} presents the proposed approach. And Section \ref{sec:setup} and \ref{sec:results} present the experimental setup and results respectively. Finally, conclusions are drawn in Section \ref{sec:conclusion}.
% Sections \ref{sec:model} to \ref{sec:results} present the proposed approach and experimental results, respectively. Finally, conclusions are drawn in Section \ref{sec:conclusion}.

%\citet{wang-etal-2020-continuity,Lee:2016:QRD:2911451.2914734} propose quotation recommendation task to help individuals know what to quote immediately in ongoing conversations. 

\section{Related Work}\label{sec:related_work}

\textbf{Quotation Recommendation.} In previous works on quotation recommendation, some efforts are made for online conversations \cite{wang-etal-2020-continuity,Lee:2016:QRD:2911451.2914734} and some for normal writing \cite{liu-etal-2019-neural-based,tan2015learning,Tan:2016:NNA:2983323.2983788}. Our work focuses on the former. For methodology, the methods they applied can be divided into generation-based framework \cite{wang-etal-2020-continuity,liu-etal-2019-neural-based} and ranking framework \cite{Lee:2016:QRD:2911451.2914734,tan2015learning,Tan:2016:NNA:2983323.2983788}. 
% \citet{tan2015learning} apply a supervised learning to rank model to integrate multiple features (e.g., author popularity, the cosine similarity relevance between quotes and contexts, etc.). \citet{Lee:2016:QRD:2911451.2914734}  present a ranking model which combines recurrent neural network and convolutional neural network.  \citet{Tan:2016:NNA:2983323.2983788} propose a neural network approach based on LSTMs. They first learn the representations for the contexts and the quotes, and then measure the relevance based on the meaning representations. \citet{liu-etal-2019-neural-based} and \citet{wang-etal-2020-continuity} model the conversations or contexts first and then generate quotes in the manner of word by word. 
Different from previous works which mainly focus on separate modeling of quotation and query and pay little attention to the relationship between them, our model directly learns the relationship between quotations and query turns based on a mapping mechanism. The relationship mapping is jointly trained with the quotation recommendation task, which improves the performance of our model.
\section{Our model}\label{sec:model}
\begin{figure}[t]
\centering
\includegraphics[width=0.45\textwidth]{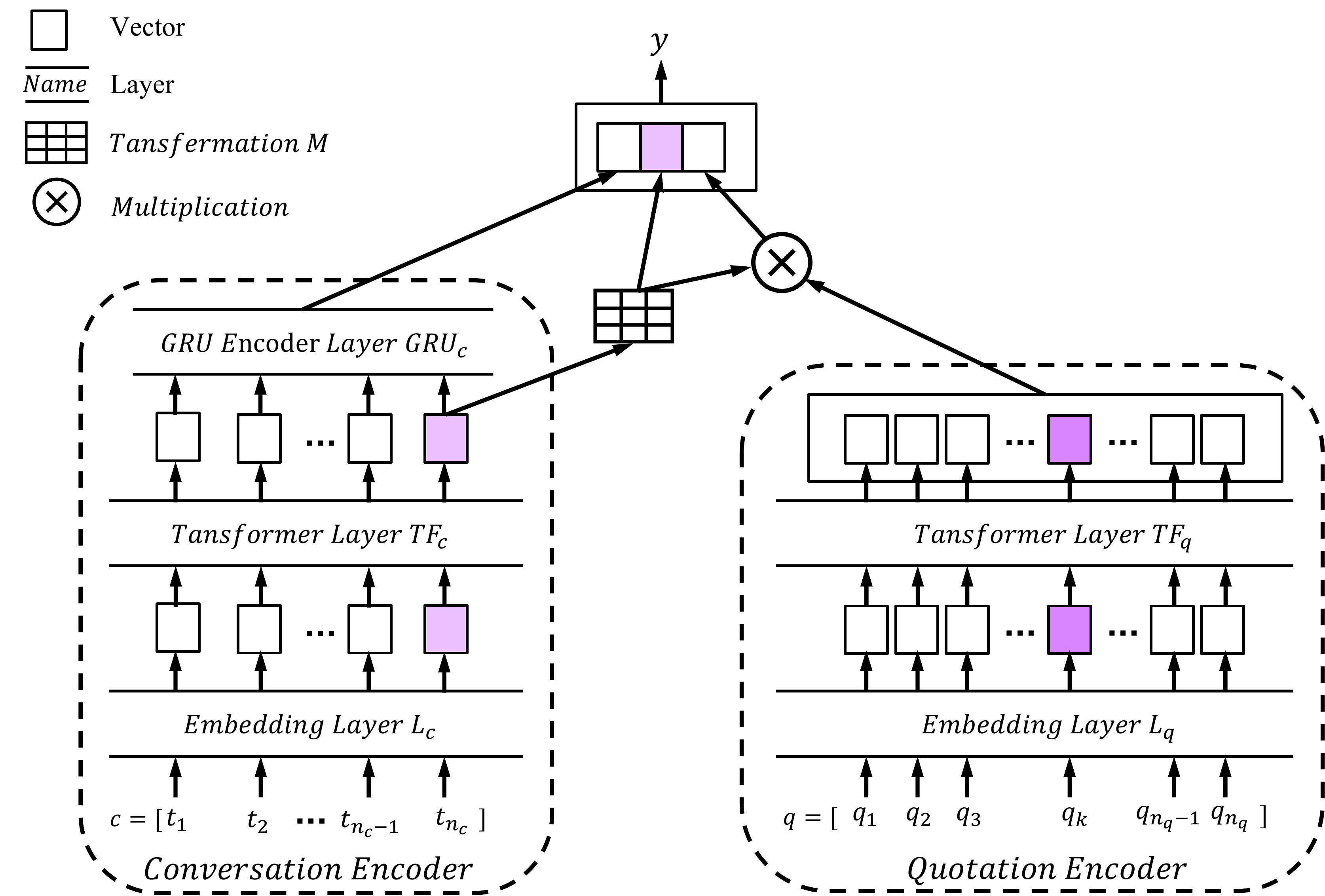}
% \vskip -0.5em
\caption{\label{fig:sketch} Our model for quotation recommendation. }
% \vskip -1.0em
\end{figure}

This section describes our quotation recommendation model, whose overall structure is shown in Figure~\ref{fig:sketch}. 
% In the following, we will first describe how we model the conversation context and query.
% Then quotation encoder is introduced, as well as the transformation operation mapping two different semantic spaces.
% Finally, we introduce our learning objectives, concerning both recommendation accuracy and transformation quality.
% 
% 
% \subsection{Problem Formulation}
The input of the model mainly contains the observed conversation $c$ and the quotation list $q$. 
%which covers all available quotations.
% and the query turns collection $h$ associated with $q$, 
% where the quotation list $q$ covers all quotations that has appeared in training corpus. 
The conversation $c$ is formalized as a sequence of turns (e.g., posts or comments) $\{t_1, t_2, ..., t_{n_c}\}$ where $n_c$ represents the length of the conversation (number of turns) and $t_{n_c}$ is the query turn. 
$t_i$ represents the $i$-th turn of the conversation and contains words $\bm{w}_i$. 
% $\bm{w}_i = \{w_{i,1}, w_{i,2}, ..., w_{i,m_i}\}$, where $w_{ij}$ is the $j$-th word in the $i$-th turn and $m_i$ is the word number of the $i$-th turn. 
The quotation list $q$ is $\{q_1, q_2, ..., q_{n_q}\}$, where $n_q$ is the number of quotations and $q_k$ is the $k$-th quotation in list $q$, containing words $\bm{w}^{\prime}_k$. 
% For any quotation $q_k$, it also contains words $\bm{w}^{\prime}_k$. 
% Noted that the words in $\bm{w}^{\prime}_k$ are usually not in the same semantic space as in $\bm{w}_i$.
Our model will output a label $y \in \{1,2,...,n_q\}$, to indicate which quotation to recommend.
% On the other hand, for any quotation $q_k$, we can extract the query turns that appeared together with it in the training set as the history queries of $q_k$, noted as $\{h_{k,1}, h_{k,2}, ..., h_{k,n_{h}}\}$, where $n_h$ is the number of history queries.

% there are some query turns $\{h_{k,1}, h_{k,2}, ..., h_{k,n_h}\}$ that associated with it, which means $h_{k,j}$ has appeared together with $q_k$ in the training set.
% We first describe the input and output for our model. The input mainly contains two parts. The first is the observed conversation $c = \{t_1, t_2, ..., t_{n_c}\}$, where $n_c$ is the turn number of $c$. Each turn $t$ is composed of several words, while the turn of the conversation, $t_{n_c}$, is the query turn that needs quotation to continue. The second part is the quotation list, where contains all the available quotations $\{q_1, q_2, ..., q_{n_q}\}$, where $n_q$ is the number of quotations. 
% Each quotation $q$ includes its words and all the history query turns that quote it: $\{h_1, h_2, ..., h_{n_h}\}$ ($n_h$ is the history number). 
% For output, we yield a Bernoulli distribution $p(c, q)$ to indicate the estimated likelihood of whether $u$ will re-engage in
% the conversation $c$, giving the chatting history $c^h$ of $u$.

\subsection{Conversation Modeling}
\label{ssec:cm}
% This subsection describes how we model the observed conversation $c$, which mainly includes two modules: turn encoder and conversation encoder. 
% \paragraph{Turn encoder.}
Our model encodes the observed conversation $c$ with a hierarchical structure, which is divided into three parts. 
The first part is an embedding layer mapping the words $\bm{w}_i$ in each turn $t_i$ into vectors. 
We then apply transformer \cite{vaswani2017attention} to learn the representation for each turn.
% since it can use self-attention mechanism to capture relative information between words and produce meaningful turn representations. 
Similar to BERT \cite{devlin2018bert}, we only use the encoder of transformer, which is stacked of several self-attention and feed-forward layers. We add a token \textsc{[CLS]} at the beginning of each turn. The hidden representation of \textsc{[CLS]} after transformer encoder is defined as the turn representation $\bm{r}^t_i$ of turn $t_i$. The procedures for the first two parts are summarized as follows:
\vskip -1em
\begin{equation}
\label{eq:tranformer}
\bm{h}_{i}^{T} = \ffn(\selfatt(\embed([w_0;\bm{w}_i])))
\end{equation}
\vskip -0.5em
\noindent where $w_{0}$ represents the \textsc{[CLS]} token, and $[;]$ indicates concatenation. Therefore $\bm{r}^t_i = \bm{h}_{i,0}^{T}$.

% \paragraph{Conversation encoder.}
% The conversation is organized as a sequence of turns ordered by replying relations. 
Next, we use a Bi-GRU \cite{cho2014learning} layer to model the whole conversation structure.
With the turn representations $\{\bm{r}^t_1, \bm{r}^t_2, ..., \bm{r}^t_{n_c}\}$  ($\bm{r}^t_{n_c}$ is the representation for the query turn) of conversation $c$ derived from previous procedure, the hidden states are updated as follows:
\vskip -1em
\begin{equation}
\label{eq:bigru}
\overrightarrow{\bm{h}}^{G}_{i} = \overrightarrow{\gru}(\overrightarrow{\bm{h}}^{G}_{i-1}, \bm{r}^t_i), \, \overleftarrow{\bm{h}}^{G}_{i} = \overleftarrow{\gru}(\overrightarrow{\bm{h}}^{G}_{i+1}, \bm{r}^t_i)
\end{equation}
\vskip -0.5em
Finally, we define the conversation representation as the concatenation of the final hidden states from two directions: $\bm{h}^{c} = [\overrightarrow{\bm{h}}^{G}_{n_c} ; \overleftarrow{\bm{h}}^{G}_{1}]$. 

\subsection{Quotation Modeling}
\label{ssec:qm}
For each quotation $q_k$ in list $q$, we extract quotation representation $\bm{r}^q_k$ with similar operation as turn representations (see Eq.~\ref{eq:tranformer}). 
% we also encode them with transformer (named as quote encoder), which contains similar structure as turn encoder.
As \citet{liu-etal-2019-neural} points out, the language used in quotations is usually different from our daily conversations, which results in two different semantic spaces. 
Therefore, we do not share the parameters of the embedding layer and transformer layers for quotations and conversation turns.
% After similar operation as Eq.~\ref{eq:tranformer}, we get the quote representation $\bm{r}^q_k$ for quotation $q_k$ ($k=1,2,..,n_q$). 
We concatenate all the quotation representations and get a combined quotation matrix $\bm{Q}$, which includes $n_q$ rows and each row represents one quotation.

\subsection{Recommendation Based on Transformation}
\label{ssec:rec}
To perform a reasonable recommendation, we consider the observed conversation $c$, the query turn $t_{n_c}$ as well as the quotation list $q$. Since they are in different semantic spaces (Section~\ref{ssec:qm}), we first map the query turns into the space of quotations with a transformation matrix $\bm{M}$. 
We assume with such transformation, the space gap can be resolved. Thus, we can calculate the distance between queries and quotations. We use $\bm{z}^c$ to represents the distances between $\bm{r}_{n_c}$ and the quotations, and it is defined with the following equation:
% we first calculate the matching degree of the query turn towards each quotation. Since the conversations and quotations are regarded as different languages and encoded with different encoders, the representations might be in different semantic space. 
% To resolve the space gap, we map the query representations into the same space of quotation, with a linear transformation $\bm{M}\in \mathbb{R}^{d \times d}$. 
% Then the matching degree $\bm{z}^c \in \mathbb{R}^{n_q}$ is derive with the following equation:

\begin{equation}
\label{eq:mapping}
\bm{z}^c = \bm{Q} \times (\bm{M} \bm{r}_{n_c})
% \vspace{-1mm}
\end{equation}
% \vskip -0.5em
% 
\noindent Finally, the output layer is defined as:
% \vskip -1.0em
\begin{equation}
\label{eq:outlayer}
\bm{y} = \bm{W} [\bm{z}^c; \bm{h}^c; \bm{M} \bm{r}_{n_c}] + \bm{b}
% \vspace{-1mm}
\end{equation}
% \vskip -0.5em
\noindent where $\bm{W}$ and $\bm{b}$ are learnable parameters. We recommend the quotations with the top $n$ highest probabilities, which are derived with a softmax function:
% Then the probability for each quotation $i$ regarding conversation $c$ can be derived from a softmax function:
% \vskip -1.0em
\begin{equation}
\label{eq:out-softmax}
p(\hat{q}=i) = \frac{\exp(y_i)}{\sum_{k=1}^{n_q} \exp(y_k)}
% \vspace{-1mm}
\end{equation}
% \vskip -4.0em
\subsection{Training Procedure}
We define our training objective as two parts. The first part is called recommendation loss, which
is the cross entropy over the whole training corpus $\mathbb{C}$:
% \vskip -1.0em
\begin{equation}
\label{eq:ce-loss}
\mathcal{L}_{rec} = - \sum_{c \in \mathbb{C}} \log \, p(\hat{q}=q^c | c,q)
% \vspace{-1mm}
\end{equation}
% \vskip -0.5em
where $q^c$ is the ground-truth quotation for conversation $c$ in training corpus. The second part is to help on the learning of transformation matrix $\bm{M}$, where we minimize the distance between the transformed query turn representation and the corresponding ground-truth quotation:
% 
% To enhance the learned representations, we also designed two additional losses to assist the training.
% 
% \paragraph{Mapping Loss.}
% This loss minimizes the difference between quotation representations and their corresponding query turns, so that the mapping between these two language spaces is more reliable. We use MSE loss to measure their difference:
% \vskip -1.0em
\begin{equation}
\label{eq:map-loss}
\mathcal{L}_{map} = \sum_{c \in \mathbb{C}}||\bm{M}\bm{r}_{n_c} -\bm{r}^q_{q^c}||_2^2
% \vspace{-1mm}
\end{equation}
% \vskip -0.5em
% where $\bm{r}^q_{q^c}$ is the representation for ground-truth quotation $q^c$. 
To train our model, the final objective is to minimize $\mathcal{L}$, the combination of the two losses:
% \vskip -1.0em
\begin{equation}
\label{eq:final-loss}
\mathcal{L} = \mathcal{L}_{rec} + \lambda \cdot \mathcal{L}_{map}
% \vspace{-1mm}
\end{equation}
% \vskip -0.5em
where $\lambda$ are the coefficient determining the contribution of the latter loss.
\section{Experimental Setup}\label{sec:setup}
\paragraph{Datasets.} We conduct experiments based on datasets from two different platforms, Weibo and Reddit, released by  \citet{wang-etal-2020-continuity}. 
% To save the space, we leave the statistics of them in Appendix.
% The statistics of these two datasets are shown in Table \ref{statistics-table1}. 
% As we can see that 
%There are around 1000 quotations for both Weibo and Reddit datasets. 
%The vocabulary size of conversations is much larger than that of quotations, while their expressing styles are also different. 
%Therefore, quotations can be viewed as a kind of different language with different word distribution compared to daily expressions. 
To make our experimental results comparable to \citet{wang-etal-2020-continuity}, we utilize their preprocessed data directly. 
\paragraph{Parameter Setting.} 
We first initialize the embedding layer with 200-dimensional Glove embedding \cite{pennington2014glove} for Reddit and Chinese words embedding \cite{song2018directional} for Weibo. For transformer layers, we choose the number of layers and heads as (2, 3) for Reddit and (4, 4) for Weibo. For the hidden dimension of transformer layers and BiGRU layers (each direction), we set it to $200$.
% For our BiGRU layers, we set the size of hidden states for each direction to $200$. 
We employ Adam optimizer~\cite{kingma:adam} with initial learning rate with 1e-4 and early stop adoption~\cite{caruana2001overfitting} in training. The batch size is set to 32.
%whose Twitter version is used for Twitter dataset and Common Crawl version is applied for Reddit\footnote{\url{https://nlp.stanford.edu/projects/glove/}}. 
Dropout strategy~\cite{Srivastava:2014:DSW:2627435.2670313} and $L_2$ regularization are used to alleviate overfitting. And the tradeoff parameter $\lambda$ is chosen from \{1e-4, 1e-3\}. All the hyper-parameters above are tuned on the validation set by grid search.

\paragraph{Evaluation and Comparisons.}
% As we formulate quotation recommendation as a multi-classification task to predict which quotation to be recommended, 
% We prefer normal evaluation metrics in multi-classification. Therefore,
Our model returns a quotation list arranged in descending order of likelihood of recommendation for each conversation. Therefore, 
we adopt MAP (Mean Average Precision), P@1 (Precison@1), P@3 (Precison@3), and nDCG@5 (normalized Discounted Cumulative Gain@5) for evaluation. 

% \paragraph{Comparisons.} 
% Here we introduce the baselines we compared. Each model we introduce below will return a quotation list that arranged in descending order of likelihood of recommendation. 
For comparison, 
% we first consider two naive baselines to show how challenge of this task, and 
we compare with previous works that focus on quotation recommendation. Below shows the details: \\
% and to understand whether our proposed model is actually learning from the data at all. 
% Apart from this, we compare with the results from \citet{wang-etal-2020-continuity} (3-) and we
% also define several other baselines () to validate the superiority of our proposed model. All of the baseline methods are listed below: \\
%1) \textsc{\underline{Random}}. The recommended quotations are arranged randomly. \\
% 1) \textsc{\underline{Frequency}}. We rank the quotations by their frequency (the time appearing in training set). \\
1) \textsc{\underline{LTR}} (Learning to Rank). We first collect features (e.g., frequency,
% cosine similarity between quotes and contexts, LDA,
Word2Vec, etc.) mentioned in \citet{tan2015learning} and then use the learning to rank tool RankLib \footnote{\url{https://github.com/danyaljj/rankLibl}} to do the recommendation. \\
% non-neural learning to rank model (henceforth \textsc{\underline{LTR}}) with handcrafted features proposed in ~\citet{tan2015learning}.
2) \textsc{\underline{CNN-LSTM}}. We implement the model proposed in \citet{Lee:2016:QRD:2911451.2914734}, which adopts CNN to learn the semantic representation of each turn and then uses LSTM to encode the conversation. \\
%4) \textsc{\underline{CNN-LSTM}}~\cite{Lee:2016:QRD:2911451.2914734}: previous quotation recommendation model (CNN for turn and quotation encoding and LSTM for conversation structure). \\
% 5) \textsc{\underline{BERT-LSTM}}. We adopt BERT~\cite{devlin2018bert} to encode each turn's text and get the turn representation, followed by an LSTM layer modeling the conversation structure. \\
3) \textsc{\underline{NCIR}}. It formulates quotation recommendation as a context-to-quote machine translation problem by using the encoder–decoder framework with attention mechanism~\cite{liu-etal-2019-neural}.  \\
4) \textsc{\underline{CTIQ}}. The SOTA model \cite{wang-etal-2020-continuity}, which employs an encoder-decoder framework enhanced by Neural Topic Model to continue the context with a quotation via language generation. \\
5) \textsc{\underline{BERT}}. We encode the conversation by BiLSTM on the BERT representations for the turns, followed by a prediction layer.
% Neural topic model is also employed to learn the topic consistency and help model the conversation and help generate words in topic.

\section{Experimental Results}\label{sec:results}
%\fcolorbox{red!20}{red!20}{Text with a red frame}
% In this section, we first introduce the main comparison results in Section \ref{ssec:exp:main_results}. 
% Then we show the effects of our methods in detail by comparing with other methods in Section \ref{ssec:comparison_in_detail}. We also show how our methods make effects in Section \ref{ssec:why} and more discussion are given in Section \ref{ssec:more_discuss}.
\begin{table}[t]\setlength{\tabcolsep}{0.4mm}\footnotesize
\newcommand{\tabincell}[2]{\begin{tabular}{@{}#1@{}}#2\end{tabular}}
\begin{center}
\scalebox{0.9}{
\begin{tabular}{lcccccccc}
\hline 
\multirow{2}{*}{Models} 
&\multicolumn{4}{c}{ \tabincell{c}{\textbf{Weibo}} }
&\multicolumn{4}{c}{ \tabincell{c}{\textbf{Reddit}} }

\\
\cline{2-9}
& MAP  & P@1 & P@3 & NG@5  & MAP  & P@1 & P@3 & NG@5  \\
\hline
\underline{\textbf{Baselines}} & & && & & & &\\
%\textsc{Random}&0.6 &0.1 &0.1&0.3&0.5 &0.1 &0.1 &0.1 \\
%\textsc{Frequency} &7.1 &2.6 &7.1 &6.6 &4.7 &1.0 &3.0 &2.9 \\
\textsc{LTR} &9.3 &3.6 &8.5 &8.1 &7.1 &1.7 &6.4 &6.2 \\
\textsc{CNN-LSTM}&11.3 &7.3 &11.0 &10.8 &5.2 &4.1 &7.0 &6.9 \\
% \textsc{SEQ2SEQ}&24.1 &19.9 &25.7 &23.2 &9.8 &7.2 &10.6 &10.1 \\
\textsc{NCIR}& 26.5 &22.6 &27.8 &26.7 &12.2 &7.3 &12.3 &11.4 \\
\textsc{CTIQ}&30.3 &27.2 &33.2 &31.6 &21.9 
&17.5 &25.8 &23.8 \\
\textsc{BERT}&31.4 &27.9 &34.0 &32.3 &26.4 &18.0 &30.2 &28.5 \\
\hline
% \underline{\textbf{Our Model}} & & &&& & & &\\
\textbf{\textsc{Our Model}}&\textbf{34.9} &\textbf{30.3} &\textbf{36.1} &\textbf{34.9} &\textbf{31.8} &\textbf{23.3} &\textbf{35.0} &\textbf{32.1} \\
% \textsc{W/O Mapping}& & &&& & & & \\
% \textsc{GRU-GRU}& & &&& & & & \\

\hline
\end{tabular}
}
\end{center}
% \vskip -1em
\caption{\label{tab:main} 
Main comparison results on Weibo and Reddit datasets (in \%). NG@5 refers to NDCG@5. The best results in each column are in \textbf{bold}. Our model yields significantly better scores than all other comparisons for all metrics ($p < 0.01$, paired t-test).
%RG-1 and RG-L refer to ROUGE-1 and ROUGE-L respectively. The best results in each column are in \textbf{bold}. Our full model \textsc{IE+QE+NTM} achieves significantly better performance than all the comparisons (paired t-test. $\ddagger$: $p<0.05$; $\dagger$: $p<0.01$)
}
% \vskip -0.5em
\end{table}
\subsection{Quotation Recommendation}
Table~\ref{tab:main} displays the recommendation results comparing our model with the baselines on Weibo and Reddit datasets. Our model achieves the best performance, exceeding the baselines by a large margin, especially on Reddit dataset. The fact that better performance comes from \textsc{BERT} and our model indicates the importance of learning efficient content representations. Our model further considers the mapping between different semantic spaces, resulting in the best performance.

% The followings are detailed observations: (1)Our model notably outperforms all the other baselines on both two datasets, which may due to our model captures the relationship batween quotations and queries directly. (2)

\paragraph{Ablation Study.}
We conduct an ablation study to examine the contributions of different modules in our model.
We replace the transformer layers with Bi-GRU (W/O Transformer) to examine the effects of different turn encoders. We also compare the models by removing transformation matrix $M$ (W/O $M$) or mapping loss $\mathcal{L}_{map}$ (W/O $\mathcal{L}_{map}$). The results are shown in Table~\ref{tab:ablation}. As can be seen, each module in our model plays a role in improving performance. The largest improvement comes from applying transformers as our encoders. The performance drop due to removing transformation and mapping loss justifies our assumption of different semantic spaces between quotations and queries.

\begin{table}[t]\setlength{\tabcolsep}{0.6mm}\small
\newcommand{\tabincell}[2]{\begin{tabular}{@{}#1@{}}#2\end{tabular}}
\begin{center}
\scalebox{0.9}{
\begin{tabular}{lcccccc}
\hline 
\multirow{2}{*}{Models} 
&\multicolumn{3}{l}{ \tabincell{l}{\textbf{Weibo}} }
&\multicolumn{3}{l}{ \tabincell{l}{\textbf{Reddit}} }
% \cmidrule(lr){2-2}\cmidrule(lr){3-3}
% 		\cmidrule(lr){4-4}\cmidrule(l){5-5}
\\
\cline{2-7}
& MAP  & P@1  & NDCG@5  & MAP  & P@1  & NDCG@5  \\
\hline
\hline
W/O Transformer &29.9  &25.9 &29.8 &25.8 &17.4 &25.7 \\
W/O $M$ &31.7  &27.4 &31.8 &29.5 &21.6 &29.5 \\
W/O $\mathcal{L}_{map}$ &32.6  &28.4 &32.4 &30.4 &22.6 &30.6 \\
Full Model& 34.9 &30.3  &34.9 &31.8 &23.3 &32.1 \\
% \textsc{W/O Transformer}& & &&& & & & \\
% \textsc{W/O Mapping}& & &&& & & & \\
% \textsc{GRU-GRU}& & &&& & & & \\

\hline
\end{tabular}
}
\end{center}
% \vskip -1em
\caption{\label{tab:ablation} Comparison results of different variants of our model on Weibo and Reddit datasets (in \%).
%RG-1 and RG-L refer to ROUGE-1 and ROUGE-L respectively. The best results in each column are in \textbf{bold}. Our full model \textsc{IE+QE+NTM} achieves significantly better performance than all the comparisons (paired t-test. $\ddagger$: $p<0.05$; $\dagger$: $p<0.01$)
}
% \vskip -1em
\end{table}

\subsection{Quotation Interpretation}
We also explore how to interpret the figurative language of quotations with our model.
% In our writing (either formal writing or online discussions), we generally decide what to quote based on the context that we are currently involved in. Intuitively, some phrases in the context may signal the important information relative to the quotation. To examine whether our model indeed learns useful knowledge and perform recommendation reasonably, 
We first extract the queries that are related to one certain quotation as history queries, then compute quotation-aware attention over all history queries. Specifically, for quotation $q_k$, with its relative history queries $\{h_{1}, h_{2}, ..., h_{m_k}\}$ from the corpus ($m_k$ is the history number), we can compute their quotation-aware attention (query-level) with their representations derived from our model :
\vskip -0.8em
\begin{equation}\small
\label{eq:hist-att}
a_{k,i} = \frac{\exp(\bm{r}^q_{k} \cdot \bm{r}_{h_{i}})}{\sum_{j=1}^{m_k} \exp(\bm{r}^q_{k} \cdot \bm{r}_{h_{j}})} 
\end{equation}
\vskip -0.2em
On the other hand, we can extract the scores for the words in each history query with their self-attention weights (word-level) in transformer.
% where the queries with the highest attention scores are defined as the most relative ones. Finally, we can extract the most relative words according to the self-attention weights from transformer encoder.
Finally, the indicative words of one quotation are those with the highest scores after the multiplication of query-level and word-level attention scores.
\begin{figure}[t]\footnotesize
    \centering
    \begin{tabular}{|p{7.2cm}|}
    \hline
    % \\
    %Queries associated with quotation \\
    \fcolorbox{green!10}{green!10}{$[h_1]$}: \fcolorbox{red!3.9}{red!3.9}{Anyone}\fcolorbox{red!0.2}{red!0.2}{that}\fcolorbox{red!38.7}{red!38.7}{spends}\fcolorbox{red!0.2}{red!0.2}{that}\fcolorbox{red!0.2}{red!0.2}{much}\fcolorbox{red!29.6}{red!29.6}{money}\fcolorbox{red!0.9}{red!0.9}{just}
    \fcolorbox{red!0.5} {red!0.5}{to}\fcolorbox{red!0.2}{red!0.2}{get}\fcolorbox{red!2.7}{red!2.7}{different}\fcolorbox{red!4.8}{red!4.8}{writing}
    \fcolorbox{red!0.8}{red!0.8}{on}\fcolorbox{red!0}{red!0}{a}\fcolorbox{red!14.8}{red!14.8}{box} ..... \\
    \fcolorbox{green!30}{green!30}{$[h_2]$}:\fcolorbox{red!4.1}{red!4.1}{And}\fcolorbox{red!0.6}{red!0.6}{that}\fcolorbox{red!1.4}{red!1.4}{'s}\fcolorbox{red!1.3}{red!1.3}{probably}\fcolorbox{red!10.3}{red!10.3}{why}\fcolorbox{red!1.1}{red!1.1}{you}\fcolorbox{red!6.4}{red!6.4}{'ll}\fcolorbox{red!13.0}{red!13.0}{never}
    \fcolorbox{red!0.1}{red!0.1}{have}\fcolorbox{red!0.1}{red!0.1}{a}\fcolorbox{red!11.0}{red!11.0}{billion}\fcolorbox{red!37}{red!37}{dollars}. \\
    \fcolorbox{green!20}{green!20}{$[h_3]$}:\fcolorbox{red!3.1}{red!3.1}{Seriously}.\fcolorbox{red!6.55}{red!6.5}{Why}\fcolorbox{red!0.4}{red!0.4}{do}\fcolorbox{red!1.2}{red!1.2}{people}\fcolorbox{red!0.3}{red!0.3}{not}\fcolorbox{red!0.4}{red!0.4}{do}\fcolorbox{red!36.7}{red!36.7}{market}
    \fcolorbox{red!13.9}{red!13.9}{research}\fcolorbox{red!17.8}{red!17.8}{before}\fcolorbox{red!16.7}{red!16.7}{buying}\fcolorbox{red!2.1}{red!2.1}{something}!?! \\
    \hline
    \fcolorbox{blue!35}{blue!35}{idiots}\fcolorbox{blue!30}{blue!30}{money}\fcolorbox{blue!25}{blue!25}{buy}\fcolorbox{blue!22}{blue!22}{pay}\fcolorbox{blue!19}{blue!19}{say}\fcolorbox{blue!16}{blue!16}{fool}\fcolorbox{blue!14}{blue!14}{alone}\fcolorbox{blue!12}{blue!12}{gamble}... \\
    %Indicative Words \\
    
    \hline
    \end{tabular}
    % \vskip -1em
    \caption{Upper part: example queries associated with the quotation ``A fool and his money are soon parted.". Lower part: 
    top $8$ indicative words with the highest weighted summed self-attention scores. Darker colors represent higher weights.
    }
    \label{fig:attention}
    % \vskip -1.0em
\end{figure}

Figure \ref{fig:attention} shows an interpretation example. We display three example queries mentioned in Figure \ref{fig:intro-example}, with both their query-level attention (green) and word-level attention (red). We can find that words like ``spends'', ``money'' and ``dollars'' are assigned higher scores since they are more related to the quotation topics . We also present the most indicative words derived from all history queries (the lower part of Figure \ref{fig:attention}). We can easily infer the meaning of the quotation with the help of indicative words like ``idiots'' and ``buy''.

\section{Conclusion}\label{sec:conclusion}
In this paper, we propose a transformation from queries to quotations to enhance a quotation recommendation model for conversations. Experiments on Weibo and Reddit datasets show the effectiveness of our model with transformation. We further explore using indicative words in history queries to interpret quotations, which shows rationality of our method. 
\section*{Acknowledgements}
The research described in this paper is partially supported by HK GRF \#14204118 and HK RSFS \#3133237. We thank the three anonymous reviewers for the insightful suggestions on various aspects of this work.

\bibliographystyle{acl_natbib}
\bibliography{anthology,acl2021}

%\appendix

\end{document}